\documentclass[letterpaper, /10 pt, conference]{ieeeconf}  % Comment this line out if you need a4paper

\IEEEoverridecommandlockouts                              % This command is only needed if 
\usepackage{multirow}
\usepackage{graphicx}
\usepackage{hyperref}
\usepackage{array}
\usepackage{makecell}
\usepackage{amsmath}
\usepackage{amssymb}
\usepackage{siunitx}
\newcolumntype{L}[1]{>{\raggedright\let\newline\\\arraybackslash\hspace{0pt}}m{#1}}
\newcolumntype{C}[1]{>{\centering\let\newline\\\arraybackslash\hspace{0pt}}m{#1}}
\newcolumntype{R}[1]{>{\raggedleft\let\newline\\\arraybackslash\hspace{0pt}}m{#1}}

\title{\LARGE \bf
Incremental Few-Shot Object Detection via Simple Fine-Tuning Approach
}

\author{Tae-Min Choi$^{1}$ and Jong-Hwan Kim$^{1\ast}$, \textit{Fellow, IEEE}% <-this % stops a space
\thanks{$^\ast$ Corresponding author \newline
\indent$^{1}$ Tae-Min Choi and Jong-Hwan Kim are with the School of Electrical Engineering,
Korea Advanced Institute of Science and Technology, Daejeon 34141, Korea
(e-mail: \href{mailto:tmchoi}{tmchoi@rit.kaist.ac.kr}; \href{mailto:jhkim}{johkim@rit.kaist.ac.kr}). \newline
\indent This work was supported by the Institute of Information \& communications Technology Planning \& Evaluation (IITP) grant funded by the Korea government (MSIT) (No. 2020-0-00440, Development of Artificial Intelligence Technology that Continuously Improves Itself as the Situation Changes in the Real World)}}

\begin{document}

\maketitle
\thispagestyle{empty}
\pagestyle{empty}

%%%%%%%%%%%%%%%%%%%%%%%%%%%%%%%%%%%%%%%%%%%%%%%%%%%%%%%%%%%%%%%%%%%%%%%%%%%%%%%%
\begin{abstract}

In this paper, we explore incremental few-shot object detection (iFSD), which incrementally learns novel classes using only a few examples without revisiting base classes. Previous iFSD works achieved the desired results by applying meta-learning. However, meta-learning approaches show insufficient performance that is difficult to apply to practical problems. In this light, we propose a simple fine-tuning-based approach, the Incremental Two-stage Fine-tuning Approach (iTFA) for iFSD, which contains three steps: 1) base training using abundant base classes with the class-agnostic box regressor, 2) separation of the RoI feature extractor and classifier into the base and novel class branches for preserving base knowledge, and 3) fine-tuning the novel branch using only a few novel class examples. We evaluate our iTFA on the real-world datasets PASCAL VOC, COCO, and LVIS. iTFA achieves competitive performance in COCO and shows a 30\% higher AP accuracy than meta-learning methods in the LVIS dataset. Experimental results show the effectiveness and applicability of our proposed method\footnote{Code is available at \url{https://github.com/TMIU/iTFA}}.

\end{abstract}

%%%%%%%%%%%%%%%%%%%%%%%%%%%%%%%%%%%%%%%%%%%%%%%%%%%%%%%%%%%%%%%%%%%%%%%%%%%%%%%%
\section{INTRODUCTION}

Deep learning using convolutional neural networks has recently made significant progress in computer vision \cite{krizhevsky2012imagenet, simonyan2014very, ren2015faster, he2017mask, goodfellow2014generative}. These works, however, are usually developed in a supervised learning scheme that relies on abundant labeled datasets. Large-scale dataset-based training requires enormous human resources for data collection and annotation. It is also challenging to adapt models to novel classes with scarce data. As a result, few-shot learning has become a key area for many computer vision researchers.

In few-shot learning, it is difficult to train a model directly from a few novel data; therefore, we train general knowledge to the model with abundant base classes and transfer it to the novel classes. Yet, most of the work is focused on image classification tasks. Few-shot object detection (FSOD) is a more challenging task because it needs to train object recognition and localization using a small training set. Several methods \cite{wang2020frustratingly, xing2019adaptive, wu2020multi, sun2021fsce, kim2020non, karlinsky2019repmet, xiao2020few, kang2019few, cao2021few} have been presented to overcome FSOD; meta-learning and fine-tuning based methods have been the main approaches to FSOD. Meta-learning studies have introduced a learning-to-learn scheme. They learn general knowledge by episodically training the model with base class data and transferring it to the novel classes. These works can adapt quickly without extra training but have shown low performance that is difficult to apply in real-world applications. Alternatively, fine-tuning methods were developed based on the two-stage fine-tuning approach (TFA) \cite{wang2020frustratingly} and outperformed meta-learning works.

Many FSOD methods usually only focus on the performance of the novel classes. Therefore, when adding the novel classes, they can not preserve the knowledge of the base classes. This leads to catastrophic forgetting. To overcome this issue, TFA fine-tunes the last layers of the object detector with a balanced dataset, including both the base and novel classes, to prevent forgetting the base classes. In reality, novel classes can be detected on the go due to the complex real-world environment. This creates problematic settings that cannot be revisited in the base classes. The FSOD methods degrade in these incremental learning settings, wherein they tend to forget base information when trained on a few novel data. In contrast, humans can learn a new concept incrementally using only small data and preserve prior knowledge without accessing the base data. Motivated by this gap between humans and machines, incremental few-shot object detection (iFSD) was proposed by \cite{perez2020incremental}. iFSD uses the same dataset as FSOD with abundant base classes and a few novel classes. Also, the training strategy of iFSD is the same as FSOD; the model is first trained with base classes, and then the detector is incrementally learned using the novel classes. However, unlike FSOD, iFSD focuses on the performance of both the base and novel classes and transfers base knowledge to the novel classes without access to base class data.

In this paper, we propose a simple fine-tuning approach for iFSD, called iTFA. Inspired by TFA, we employ a two-stage training strategy based on Faster R-CNN \cite{ren2015faster}. We first train the Faster R-CNN with abundant base classes and then fine-tune the RoI feature extractor and classifier using a few novel classes. We focus on fine-tuning layer decisions and model design by analyzing the structure of Faster R-CNN and exploring the compatibility of the classifier optimized for iFSD.

TFA \cite{wang2020frustratingly} prevented catastrophic forgetting due to a balanced fine-tuning training set, including the base classes. However, as mentioned above, we only access a few novel classes after training with the base classes in iFSD. Therefore, the two-stage training scheme must be modified to be suitable for iFSD. We train the object detector with a class-agnostic regressor using abundant base class data. The RoI feature extractor and classifier are separated and connected parallel to the base and novel branches. We also explore the effect of fine-tuning layer decision of the RoI feature extractor and examine the compatibility of the classifier. With these modifications, iTFA outperforms the baselines on the novel class accuracy and overcomes catastrophic forgetting. In this paper, iTFA is evaluated on the PASCAL VOC \cite{everingham2010pascal}, COCO \cite{lin2014microsoft}, and LVIS \cite{gupta2019lvis} datasets for the iFSD task. Experiments show the effectiveness and competitive performance of our iTFA in all datasets.

To summarize, The contribution of this paper is threefold: (1) We propose the Incremental Two-stage Fine-tuning Approach (iTFA), which is a simple fine-tuning strategy optimized for iFSD; (2) Concretely, we leverage a class-agnostic regressor trained on large-scale data. We also separate the base and novel branches to prevent catastrophic forgetting and empirically explore compatibility between the fine-tuning layer and classifier for the accurate performance on novel classes; (3) Extensive experiments on the iFSD benchmark demonstrate higher performance on novel classes and prevent degradation of the base class accuracy.

\section{RELATED WORK}

\subsection{Few-shot object detection}
FSOD aims to detect novel classes with a few labeled novel class samples and abundant base class samples. Many early works on FSOD apply meta-learning \cite{kang2019few, yan2019meta, xiao2020few, karlinsky2019repmet}, which uses the base samples to train a meta-learner to class agnostic knowledge so that they can be adapted to novel classes without extra training. Recently, \cite{wang2020frustratingly} introduced TFA, which outperforms meta-learning-based methods. They presented a simple training strategy that fine-tunes only the last layers of the object detector. Following this approach, various fine-tuning-based methods \cite{sun2021fsce, wu2020multi, cao2021few, xing2019adaptive} have achieved state-of-the-art performance. Inspired by this workflow, we modify the TFA as a fine-tuning-based method for iFSD benchmarks.

\subsection{Incremental few-shot object detection}
iFSD was first proposed in \cite{perez2020incremental}. Unlike FSOD, iFSD does not access the base class images when learning novel classes. Due to this assumption, catastrophic forgetting occurs; the model adds novel classes, then performance degradation occurs to the base classes. Like the FSOD workflow, prior works \cite{perez2020incremental, yin2022sylph} have developed meta-learning-based methods to solve these problems. ONCE \cite{perez2020incremental} presented a class code generator to generate classification heads for novel classes. Sylph \cite{yin2022sylph} developed a code predictor head and a code process module to solve iFSD without additional training and forgetting the base classes. Meta-learning-based methods can be directly adapted to novel classes; however, they have low accuracy. Better performance is needed for its application in a practical scenario. Fine-tuning-based works \cite{li2021class, li2021towards} have been presented to achieve more accurate performance. They leverage knowledge distillation loss to overcome catastrophic forgetting. However, their network structure and training methods do not preserve the base class performance. To achieve higher accuracy in both the base and novel classes, we propose iTFA. This simple fine-tuning approach is suitable for iFSD by modifying the structure and training strategy of TFA.

\section{METHOD}

\subsection{Formulation}
\label{sec:sec3.1}
\begin{figure*}
    \centering
    \includegraphics[width=\textwidth]{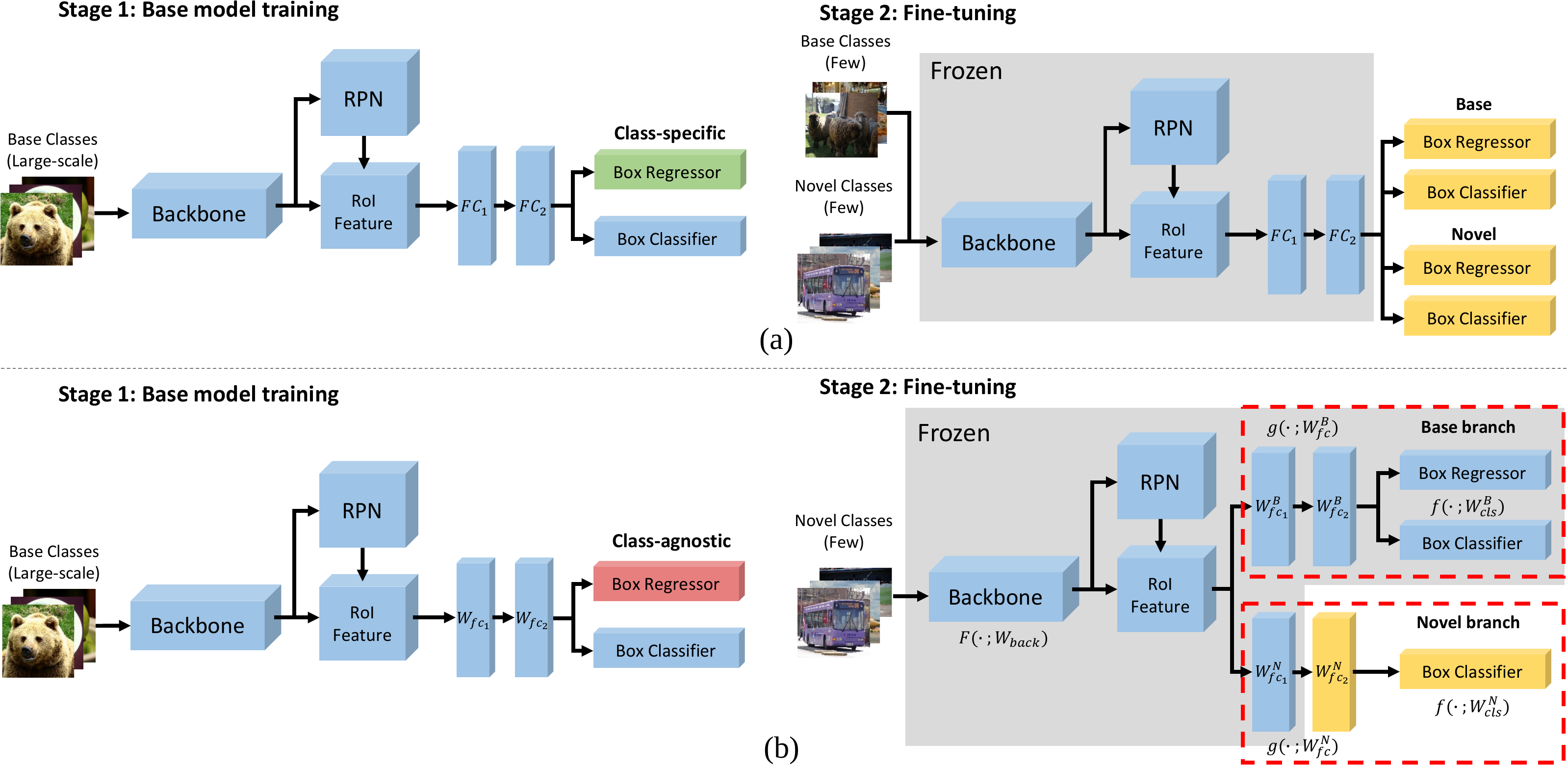}
    \caption{Method overview. (a) the illustration of TFA, (b) the overall framework of iTFA. Both frameworks consist of two stages: base model training and fine-tuning stages. The whole object detector is trained with large-scale base classes in the base model training stage. In the fine-tuning stage, each method fixes the components indicated by gray boxes and fine-tunes other structures. TFA uses both base and novel classes in the fine-tuning stage. However, iTFA only uses novel classes, which causes catastrophic forgetting. To overcome catastrophic forgetting, iTFA divides the RoI feature extractor and classifier into two branches and fine-tunes only the novel branch. Also, the class-agnostic regressor is used for easier adaptation to novel classes.}\label{fig:fig1}
\end{figure*}

We follow the iFSD settings introduced in \cite{wang2020frustratingly, yin2022sylph}. In iFSD, there are two sets of dataset with base classes, $C_{base}$, and novel classes, $C_{novel}$. Base classes have abundant annotated training data, and novel classes have few training data (usually 1$\sim$10 per category). For an object detection dataset, $D =\{(x,\,y),\, \,x\in X,\,y\in Y\}$, where $x$ is the input image and $y = \{(c_i,\,b_i),\, i=1,\ \ldots,\ N\}$ denotes the categories $c$ and bounding box coordinates $b$ of the $N$ object instances in the image $x$. The goal of iFSD is to incrementally train the object detector to recognize both novel and base classes without revisiting base classes. As shown in Fig. \ref{fig:fig1}, we split the training stage into the \textit{base model training stage} and \textit{fine-tuning stage}. We first pre-train the object detector with abundant base classes in the base model training stage. In this stage, we use $D_{base}$ containing only base classes $c \in C_{base}$ with sufficient labeled images. After the first stage, we transfer prior knowledge to classify the novel classes that can use only $K$ shots training data while maintaining the performance of base classes. In the fine-tuning stage, we use $D_{novel}$ containing novel classes $c \in C_{novel}$ with $K$ examples per class. In the test stage, our work aims to train a model that performs well on both base and novel classes, so we use $D_{test}$ that contains all classes $c \in (C_{base} \cup C_{novel}$) at the test stage.

\subsection{Incremental two-stage fine-tuning approach}
\label{sec:sec3.2}
In this section, we introduce our iTFA method for iFSD. As shown on the right side of Fig. \ref{fig:fig1}(a), TFA uses base classes in the fine-tuning stage to prevent forgetting the prior knowledge. However, in iFSD, the object detector is fine-tuned only using novel classes without access to the base classes. TFA also freezes all structures before the box classifier, which disturbs the feature alignment of novel classes in the iFSD setting. To overcome these limitations, we propose iTFA, a simple fine-tuning approach optimized for iFSD. The training strategy of our iTFA has three steps: 1) changing the class-specific regressor to the class-agnostic regressor in the base training stage, 2) separating the RoI feature extractor (composed of two fc layers) and the classifier of Faster R-CNN \cite{ren2015faster} into the novel and base class branches, and 3) analyzing the effect of the decision of fine-tuning layers and compatibility of the classifier, then fine-tuning the novel branch.

% The main difference between the problem dealt with in TFA \cite{wang2020frustratingly} and iFSD is whether or not the base class is included in the training set of the fine-tuning stage. Figure 1(a) shows that a balanced training set, including base and novel classes, is created to prevent forgetting the base class in TFA. However, in iFSD, we incrementally train the object detector only using novel classes without access to the base classes. Also, TFA freezes all structures before the classifier, which disturbs the feature alignment of novel classes. Therefore, the training strategy proposed by TFA is not optimized for iFSD. We propose iTFA, a fine-tuning strategy optimized for iFSD. The key component of our method is as follows. 1) separate the RoI feature extractor (composed of two fc layers) and classifier of Faster R-CNN \cite{ren2015faster} into the novel and base class branches, 2) analyze the effect of fine-tuning each fc layer and fine-tune only novel branch, and 3) change the class-specific regressor to class-agnostic regressor. Our method provides a baseline of a fine-tuning-based method optimized for iFSD while using Faster R-CNN.

% We experimented with classifier and box regressor structures suitable for iFSD to achieve increased novel class performance.
\subsubsection{Base model training} 
In the base training stage, we train the feature extractor, the classifier, and the box regressor using the base classes dataset $D_{base}$ (left side of Fig. \ref{fig:fig1}). We use the same training strategy from Faster R-CNN \cite{ren2015faster}. The loss term for training is as follows:
\begin{equation}
\label{eqn:eq1}
    L = L_{rpn} + L_{cls} + L_{loc},
\end{equation}
where $L_{rpn}$ is leveraged to determine whether the output of Region Proposal Network (RPN) is foreground or background and to adjust the anchor position, $L_{cls}$ is the cross-entropy loss of the classifier, and $L_{loc}$ is the $L_1$ loss for the box regressor.

\subsubsection{Class-agnostic box regressor} 
Previous iFSD methods \cite{perez2020incremental, wang2020frustratingly, li2021towards, li2021class} applied a class-specific box regressor to the object detector. A class-specific box regressor can generate a bounding box fitted to each class. However, it is challenging to train a class-specific regressor in few-shot learning due to the few instances of novel classes. It is difficult to obtain various bounding box information with a small number of instances because even the same class has different bounding box characteristics depending on the style and composition of the image. Therefore, we change the class-specific regressor to the class-agnostic box regressor in the base training stage and use it to predict the bounding boxes of both base and novel classes. The class-agnostic regressor can be leveraged for the novel classes because it has been pre-trained on the abundant dataset. When in the fine-tuning stage, the regressor is frozen, so it does not overfit to novel classes and keeps the knowledge of the base classes. Also, by freezing the class-agnostic box regressor, we can focus on classification rather than localization and prevent catastrophic forgetting of the base classes. 

\subsubsection{Fine-tuning} 
In the fine-tuning stage, the novel classes are trained incrementally with $D_{novel}$ ($K$ shots per class) without access to $D_{base}$ (right side of Fig. \ref{fig:fig1}(b)). Only the $L_{cls}$ in (1) is used due to freezing the backbone, RPN, and class-agnostic regressor. In this stage, our goal is to maintain the performance of base classes and adapt to the novel classes with a few images. In the following, we specify our fine-tuning strategy to achieve these goals.

% This is a different condition from TFA, which constructs a balanced dataset using a few images of base classes. We tried to maintain the performance of base classes and improve the accuracy of novel classes by modifying three parts of TFA.
%%%%%%%%%%%%%%%%%%%%%%%%%%%%%%%%%%%%%%%%%%%%%%%%%%%%%%%%%%%

\subsubsection{Separate novel branch}
TFA separates feature representation learning and box classifier learning into two stages. This means feature representations from the base classes could be transferred to novel classes without an extra weights update. The base model trained with sufficient base class data has a well-aligned feature space. Moreover, the classifier, which is fine-tuned with a few shot data including both $C_{base}$ and $C_{novel}$, classifies novel classes without forgetting the decision boundary of the base classes. However, in iFSD, we cannot use $C_{base}$ during fine-tuning. If the base classifier is fine-tuned without base class data, it will cause overfitting on the novel classes and lead to catastrophic forgetting of the base classes. Therefore, we first separate the base and novel classifiers and then freeze the base classifier and fine-tune only the novel classifier. Separating the novel and base classifiers allows us to focus on feature alignment for novel classes only. For this purpose, as shown in Fig. \ref{fig:fig1}(b), we add the RoI feature extractor ($W_{fc_1}^N$ and $W_{fc_2}^N$) for the novel classes in parallel, and it is trained to align the novel class features. We only separate the RoI feature extractor because the RoI features generated from the backbone and RPN have sufficient information about the objects in the image due to being trained with abundant base classes. 

We design the fine-tuning strategy to align the novel feature space as follows:
\begin{gather}
    \label{eqn:eq2}
    \min_{W_{fc_2}^{N}, W_{cls}^{N}} L_{cls}(y_i,\ [p^B_i,\ p^N_i]),\\ 
    \nonumber with \ p^B_i = f(g(z_i;\ W_{fc}^{B});\ W_{cls}^{B}),\\
    \nonumber p^N_i = f(g(z_i;\ W_{fc}^{N});\ W_{cls}^{N}), z_i = \mathcal{F}(x_i;\ W_{back}),    
\end{gather}
where $f(\cdot\ ;\ W_{cls}^B)$ and $f(\cdot\ ;\ W_{cls}^N)$ are the base and novel classifiers, respectively. $g(\cdot \ ;\ W_{fc}^B)$ and $g(\cdot \ ;\ W_{fc}^N)$ are the RoI feature extractors for the base and novel classes, respectively. $W_{fc_2}^N$ is the last fc layer of the novel RoI feature extractor. $W_{back}$ is the backbone network, and $z_i$ means representation vector created from the backbone. $p^B_i$ is ($\lvert C_{base}\rvert$+1)-way including the background as base classes prediction, and $p^N_i$ is $\lvert C_{novel}\rvert$-way as novel classes prediction. Then, $L_{cls}$ is calculated by concatenating the two predictions. $W_{cls}^N$ is randomly initialized and pre-trained weights $W_{fc_1}$ and $W_{fc_2}$ from the base training stage are loaded into $W_{fc_1}^N$ and $W_{fc_2}^N$, respectively. Other networks are frozen except $W_{cls}^N$ and $W_{fc_2}^N$ to preserve their prior knowledge. In particular, as shown in Fig. \ref{fig:fig1}(b), we fine-tune only $W_{fc_2}^N$ in the fine-tuning stage because, empirically, fine-tuning $W_{fc_1}^N$ is found to disturb the disentangling of novel features. The effect of fine-tuning layers is evaluated in Section \ref{sec:abl}.
%  Also, We separate the classifiers and RoI feature extractors ($W_{fc}^B$ and $W_{fc}^N$) parallel to the RoI features for disentangling base and novel classes. 

\subsubsection{Compatibility of classifier} 
Since the feature spaces of the base and novel classes are separated, we consider the compatibility of each RoI feature and classifier. In few-shot learning, there is a bias in which classifiers predict base classes with higher scores, which makes it more challenging to classify novel classes. Therefore, prior few-shot learning works use cosine classifiers \cite{gidaris2018dynamic, qi2018low, chen2019closer} with normalized features and various margin losses \cite{deng2019arcface, wang2018cosface} to align intra-class and inter-class distributions of the novel classes. Cosine similarity-based classifiers are effective for FSOD since they focus on disentangling only the feature space of the novel class. However, in iFSD, we need to align the novel class features without disturbing the feature space of the base classes. In our method, making enough space for the novel classes on the hypersphere is difficult due to freezing the base class feature space during the fine-tuning stage. The inter-class separability is violated because there is no realignment on the base class feature space. Thus, we test the compatibility of the classifier and empirically find that a linear classifier helps disentangle the novel class feature and improve the accuracy of novel classes.

\section{EXPERIMENTS}

\label{sec:exp}
\subsection{Implementation details}
\label{sec:exp4.1}
\subsubsection{Datasets and metric}
We evaluated iTFA on PASCAL VOC 2012 \cite{everingham2010pascal}, COCO \cite{lin2014microsoft}, and LVIS v1 \cite{gupta2019lvis} datasets. The evaluation metric and the data split strategy were followed from \cite{wang2020frustratingly, perez2020incremental, yin2022sylph} for a fair comparison. For COCO, 60 categories disjoint from PASCAL VOC are considered base classes; the other 20 classes are considered novel classes. For LVIS v1, 1203 categories are composed of 337 rare, 461 common, and 405 frequent classes. The 337 rare categories were treated as novel classes, and the 866 common and frequent categories as base classes. For the cross-dataset evaluation from COCO to PASCAL VOC, we used the same base and novel class set (60 and 20  categories from COCO) for the base model training and fine-tuning stage, respectively. We then evaluated the object detector on the PASCAL VOC 2012 test set. We evaluated our work on 1, 5, and 10 shots and used AP50 and the standard COCO-style metric, which calculates the average precision (AP) at multiple intersection-over-union (IoU) thresholds from 0.5 to 0.95. For stable results, ten tests were made on each random seed, and the mean average precision was reported for all experiments. For COCO, base, novel, and harmonic mean AP (mAP) were reported as bAP, nAP, and hAP, respectively. For LVIS, APr, APc, and APf are the mAP of rare, common, and frequent categories, respectively. Also, APs, APm, and APl are the mAP of small, medium, and large objects. For cross-dataset evaluation, AP50 was used for comparison to the other methods.

\subsubsection{Training setting}
Our method was implemented using MMDetection \cite{mmdetection}. We used Faster R-CNN \cite{ren2015faster} with Feature Pyramid Network \cite{lin2017feature} for all experiments. The ResNet-50/101 \cite{he2016deep} backbones were used for COCO, while ResNet-50 and ResNet-101 were used for LVIS and Pascal VOC for the fair comparison, respectively. SGD with a momentum of 0.9, a weight decay of $1\mathrm{e}{-4}$, and a batch size of 8 was used on 4 NVIDIA RTX 2080Ti GPUs, with two images per GPU. We set the learning rate to $1\mathrm{e}{-2}$ for both the base training and the fine-tuning stages. For COCO and PASCAL VOC, we trained the base model for 12 epochs and fine-tuned the object detector for 3000 steps. The learning rate decreased ten-fold at epochs 8 and 11 in the base training and at steps 1500 and 2000 in the fine-tuning stage. For LVIS, we trained the base model for 24 epochs, and the learning rate decreased at epochs 16 and 22. We fine-tuned the model for 10000 iterations, and the learning rate decreased at 7000 and 9000 iterations.

\begin{table*}[ht!]
\caption{Ablation study of fine-tuning layer decision, classifier, and regressor of iTFA. \textbf{Bold} and \underline{underline} indicate the best and the second best performance, respectively.} \label{tab:tab1}
\renewcommand{\arraystretch}{1.1}
\centering
{\small
% \resizebox{\columnwidth}{!}{%
\begin{tabular}{ccc|cc|cc|C{1cm}C{1cm}C{1cm}}
\Xhline{3\arrayrulewidth}
\multicolumn{3}{c|}{Fine-tuning layer} &
  \multicolumn{2}{c|}{Regressor} &
  \multicolumn{2}{c|}{Classifier} &
  \multicolumn{3}{c}{Shots} \\ \hline
$None$       & $fc_1$        & $fc_2$        & specific   & agnostic   & Linear     & Cosine     & 1   & 5   & 10   \\ \Xhline{2\arrayrulewidth}
\checkmark &            &            &            & \checkmark & \checkmark &            & 3.4 & 7.0 & 8.3  \\
           &            & \checkmark &            & \checkmark & \checkmark &            & \textbf{4.3} & \textbf{9.9} & \textbf{11.8} \\
           & \checkmark & \checkmark &            & \checkmark & \checkmark &            & \underline{3.6} & \underline{9.0} & 10.9 \\ \hline
\checkmark &            &            &            & \checkmark &            & \checkmark & \underline{3.6} & 7.6 & 9.2  \\
           &            & \checkmark &            & \checkmark &            & \checkmark & 3.3 & 8.9 & \underline{11.0} \\
           & \checkmark & \checkmark &            & \checkmark &            & \checkmark & 2.6 & 8.4 & 10.5 \\ \hline
           &            & \checkmark & \checkmark &            & \checkmark &            & 2.8 & 7.5 & 9.7  \\
           &            & \checkmark & \checkmark &            &            & \checkmark & 3.0 & 8.2 & 10.2 \\ \Xhline{3\arrayrulewidth}
\end{tabular}%
}
\end{table*}

\begin{table*}[ht!]
\centering
\caption{Performance of novel and base classes on the COCO dataset.} \label{tab:tab2}
\renewcommand{\arraystretch}{1.1}
{\small
% \resizebox{\columnwidth}{!}{%
\begin{tabular}{c|l|c|C{1cm}C{1cm}|C{1cm}C{1cm}|C{1cm}C{1cm}}
\Xhline{3\arrayrulewidth}
\multirow{2}{*}{Shots} & \multicolumn{1}{l|}{\multirow{2}{*}{Method}} & \multicolumn{1}{c|}{\multirow{2}{*}{Backbone}} & \multicolumn{2}{c|}{Novel} & \multicolumn{2}{c|}{Base} & \multicolumn{2}{c}{Harmonic} \\ 
                    & \multicolumn{1}{l|}{} & \multicolumn{1}{c|}{} & nAP  & nAP50 & bAP   & bAP50 & hAP   & hAP50 \\ \Xhline{2\arrayrulewidth}
\multirow{7}{*}{1}  & ONCE \cite{perez2020incremental}   &  \multicolumn{1}{c|}{\multirow{3}{*}{ResNet-50}}             & 0.7  & -     & 17.9  & -     & 1.4  & -     \\
                    & Sylph \cite{yin2022sylph}         &        & 1.1  & -     & \textbf{37.6}  & -     & 2.1  & -     \\
                    & Ours    &              & 3.8  & 6.5   & 35.7  & 55.9  & 6.8   & 11.7  \\ \cline{2-9} 
                    & TFA \cite{wang2020frustratingly}   &   \multicolumn{1}{c|}{\multirow{4}{*}{ResNet-101}}             & 1.9  & 3.8   & 31.9  & 51.8  & 3.6  & 7.1  \\
                    & iMTFA \cite{ganea2021incremental} &                & 3.2  & 5.9   & 27.8  & 40.1  & 5.8   & 10.3  \\
                    & LEAST \cite{li2021class}         &        & \underline{4.2}  & -   & 29.5  & -  & \underline{7.4}   & -  \\
                    & Ours    &              & \textbf{4.3}  & \textbf{7.2}   & \underline{37.4}  & \textbf{57.4}  & \textbf{7.7}   & \textbf{12.8}  \\ \Xhline{3\arrayrulewidth}
\multirow{7}{*}{5}  & ONCE \cite{perez2020incremental}  &   \multicolumn{1}{c|}{\multirow{3}{*}{ResNet-50}}             & 1.0  & -     & 17.9  & -     & 1.9   & -     \\
                    & Sylph \cite{yin2022sylph}       &          & 1.5  & -     & \textbf{42.4}  & -     & 2.9   & -     \\
                    & Ours      &            & 8.3 & 15.7 & 35.5 & 56.0 & 13.5 & 24.5 \\ \cline{2-9} 
                    & TFA \cite{wang2020frustratingly} &     \multicolumn{1}{c|}{\multirow{4}{*}{ResNet-101}}             & 7.0  & 13.3  & 32.3  & 50.5  & 11.5  & 21.1  \\
                    & iMTFA \cite{ganea2021incremental} &                & 6.1  & 11.2  & 24.1  & 33.7  & 9.7   & 16.8  \\
                    & LEAST \cite{li2021class}&                 & \underline{9.3}  & -  & 31.3  & -  & \underline{14.3}   & -  \\
                    & Ours      &            & \textbf{9.9} & \textbf{17.8} & \underline{37.2} & \textbf{57.3} & \textbf{15.6} & \textbf{27.2} \\ \Xhline{3\arrayrulewidth}
\multirow{8}{*}{10} & ONCE \cite{perez2020incremental}   &   \multicolumn{1}{c|}{\multirow{3}{*}{ResNet-50}}            & 1.2  & -     & 17.9  & -     & 2.3   & -     \\
                    & Sylph \cite{yin2022sylph}  &               & 1.7  & -     & \textbf{42.8}  & -     & 3.3   & -     \\
                    & Ours   &               & 10.2 & 19.3  & 35.5  & 56.0  & 15.9  & 28.7 \\ \cline{2-9} 
                    & TFA \cite{wang2020frustratingly}  &    \multicolumn{1}{c|}{\multirow{5}{*}{ResNet-101}}             & 9.1  & 17.1  & 32.4  & 50.6  & 14.2  & 25.6  \\
                    & iMTFA \cite{ganea2021incremental}  &               & 7.0  & 12.7  & 23.4  & 32.4  & 10.7  & 18.3  \\
                    & DBF \cite{li2021towards}  &                & 9.1 & -       & 28.5  &   -   &   13.8    &   -   \\
                    & LEAST \cite{li2021class}  &                & \textbf{12.8} & -       & 31.3  &   -   &   \textbf{18.2}    &   -   \\
                    & Ours   &               & \underline{11.8} & \textbf{21.3}  & \underline{37.2}  & \textbf{57.3}  & \underline{17.9}  & \textbf{31.1}  \\
\Xhline{3\arrayrulewidth}
\end{tabular}%
}
\end{table*}

\subsection{Ablation study}
\label{sec:abl}
We evaluated the following ablations on COCO to analyze how each component of iTFA affects the novel class performance. All training settings were maintained as mentioned in Section \ref{sec:exp4.1}. The performance was evaluated by changing three components: the fine-tuning layer selection, regressor, and classifier. The fine-tuning layer decides which layers of the RoI feature extractor to fine-tune. $None$ means that all layers are frozen, and only the classifier is updated. $fc_1$ and $fc_2$ indicate the fine-tuning $W_{fc_1}^N$ and $W_{fc_2}^N$, respectively. In addition, we tested the effects of class-specific and class-agnostic regressors. The effect of the classifier was tested using the linear and cosine classifier. The scale factor of the cosine classifier was set to 20.

As Table \ref{tab:tab1} shows, fine-tuning only $W_{fc_2}^N$ with a linear classifier and class-agnostic regressor outperforms other settings in all shots. This demonstrates that fine-tuning $W_{fc_2}^N$ with freezing $W_{fc_1}^N$ helps disentangle the novel class features. Also, a cosine classifier with normalized features disturbs the novel feature alignment. Compared to rows 2, 5, and the last two rows of Table \ref{tab:tab1}, Faster R-CNN with a class-agnostic regressor outperforms a class-specific regressor in all shots. This justifies that the class-agnostic regressor can be leveraged to the novel classes and is more suitable for iFSD tasks.

% We first evaluate effect of fine-tuning layer in the RoI feature extractor. Row 1, 2, and 3 of Table 1 mean the difference in fine-tuning layer of the RoI feature extractor ($W_{fc_1}^N$ and $W_{fc_2}^N$) of novel branch. 'None' means freeze both layers of RoI feature extractor. $fc_1$ and $fc_2$ mean fine-tuning $W_{fc_1}^N$ and $W_{fc_2}^N$, respectively.

% except the structure of classifier and regressor and the fine-tuning layer of the RoI feature extractor. We test linear classifier and cosine classifier with scale factor of the cosine classifier is 20 as in \cite{wang2020frustratingly}. From Table 1, when only $FC_2$ is fine-tuned with a linear classifier, it outperformed other ablations in all shots. Concretely, fine-tuning by dividing the base and novel branches performs better than fine-tuning only the classifier. This demonstrates that separating RoI feature extractors is more suitable for iFSD. Also, in most settings, the linear classifier detects novel classes more accurately than the cosine classifier. This means that normalized features disturb novel feature alignment.

\subsection{Comparison with prior works}
\subsubsection{Baselines}
We considered six baselines in our experiments. TFA \cite{wang2020frustratingly} is a fine-tuning-based approach for FSOD. ONCE \cite{perez2020incremental} and Sylph \cite{yin2022sylph} are the meta-learning-based approaches for iFSD; they use a one-stage object detector due to its superior generalization performance. iMTFA \cite{ganea2021incremental} is designed for incremental few-shot semantic segmentation based on Mask R-CNN \cite{he2017mask}; it creates novel classifier weights through weight imprinting. DBF \cite{li2021towards}, and LEAST \cite{li2021class} use Faster R-CNN as fine-tuning-based iFSD methods. DBF presents a progressive model updating rule to preserve the long-term memory of old classes. LEAST stores a few exemplars from the base training set to preserve prior knowledge when fine-tuning the detector.

\subsubsection{Results on COCO} 
As shown in Table \ref{tab:tab2}, our iTFA achieves competitive performance in all settings. In particular, iTFA outperforms TFA by 20-40\% on 5 and 10-shots and even doubles the $nAP$ on a 1-shot setting. This demonstrates that iTFA is a more suitable fine-tuning approach for iFSD tasks. In addition, our method surpasses meta-learning methods by a large margin in all environments. This corroborates that our training strategy can be applied to practical iFSD tasks. Our method also shows consistent base class performance in all shots, demonstrating that catastrophic forgetting is prevented by disentangling the feature space of the base and novel classes by separating the RoI feature extractor and classifier. Sylph shows the highest $bAP$ in all settings. This performance difference is due to the type of object detector. Sylph used the FCOS \cite{tian2019fcos} as the base detector, which outperforms Faster R-CNN. In this paper, our objective is to propose a fine-tuning strategy for the iFSD task, so we used Faster R-CNN, the most popular two-stage object detector, as the base detector. In addition, although we did not use the base class exemplars in the fine-tuning stage, we achieved more accurate performance in the 1-shot and 5-shot settings and competitive performance compared to LEAST in the 10-shot setting. LEAST shows a higher nAP at the 10-shot. LEAST stores $k$ base classes exemplars (in $k$-shot) and uses them for fine-tuning to keep the intra-class variance of base classes. It helps distinguish novel and base classes in the fine-tuning stage, so LEAST performs better in a 10-shot setting that can store many base exemplars. However, saving the base class exemplars is not a fair comparison because it violates the iFSD assumption. Despite using a basic detector and only novel class data for fine-tuning, iTFA shows almost the best performance for several environments. These results demonstrate the strength of our iTFA in preserving knowledge and accurate adaptation with a few samples.

\subsubsection{Results on LVIS v1}
Our iTFA was also evaluated on the LVIS dataset. LVIS is a natural long-tail distribution dataset with a large number of categories. Since rare classes of LVIS have 1 to 10 instances, each class is sampled uniformly with 10 instances to balance the novel set. Table \ref{tab:tab3} shows our results compared to recent works on LVIS. We tested iTFA on the iFSD setting that treated rare classes ($K<$ 10) to a novel set. Our method significantly outperforms meta-learning methods with about +4 and +12 AP gains on rare classes. Although iTFA uses only novel classes in the fine-tuning stage, it showed competitive performance in all categories against TFA and the joint-train model. This demonstrates the effectiveness of iTFA on the long-tail distribution dataset.
\vspace{-0.1cm}
\begin{table}[hbt!]
\renewcommand{\arraystretch}{1.1}
\caption{Performance of rare, common, and frequent classes on the LVIS v1 dataset.} \label{tab:tab3}
\centering
\resizebox{\columnwidth}{!}{%
{\footnotesize
\begin{tabular}{l|c|ccc|ccc}
\Xhline{3\arrayrulewidth}
Method      & AP   & APs  & APm  & APl  & APr  & APc  & APf  \\ \Xhline{2\arrayrulewidth}
Joint-train \cite{gupta2019lvis} & 23.5    & 18.2    & 30.9    & 35.4    & 10.5    & \underline{22.2}    & \textbf{30.6}    \\
TFA \cite{wang2020frustratingly}         & \textbf{24.4} & \textbf{19.9} & 29.5 & \textbf{38.2} & \underline{16.9} & \textbf{24.3} & 27.7 \\
ONCE \cite{perez2020incremental}        & 12.9 & -    & -    & -    & 6.3  & 11.2 & 17.7 \\
Sylph \cite{yin2022sylph}       & 20.7 & -    & -    & -    & 13.9 & 19.0 & 25.5 \\
Ours        & \underline{24.0} & 17.5 & \textbf{31.2} & 35.9 & \textbf{18.1} & 21.0 & \underline{29.9} \\ \Xhline{3\arrayrulewidth}
\end{tabular}%
}}

\end{table}

\subsubsection{Results on cross-domain from COCO to PASCAL VOC}
We evaluated iTFA in the cross-dataset setting from COCO to PASCAL VOC. Since PASCAL VOC has no base classes, we only report the novel class performance. Table \ref{tab:tab4} shows the same relative results evaluated on COCO. Our method performs well against the prior methods in all the settings, indicating the strength of our method.

\subsubsection{Qualitative results} 
\begin{figure}[hbt!]
    \centering
    \includegraphics[width=\columnwidth]{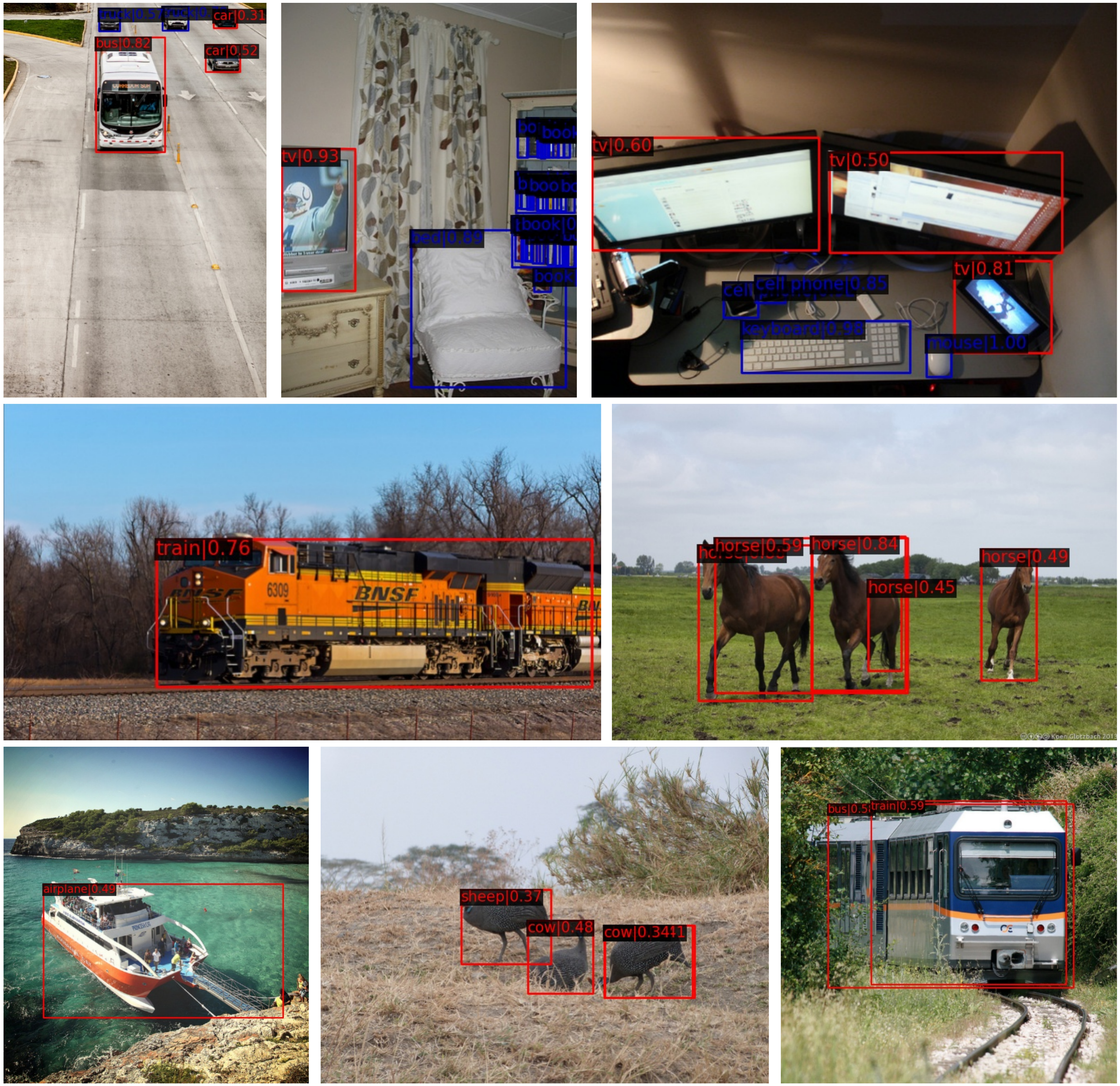}
    \caption{Qualitative results of iTFA on COCO with $K$=10. The top two rows are success cases, and the bottom row images are failure cases. The blue and red boxes indicate the base and novel classes, respectively (zoom view better).}
    \label{fig:fig2}
\end{figure}

Fig. \ref{fig:fig2} illustrates the qualitative results of our method on the COCO dataset. The top two rows show success results, and the bottom row shows failure results. The success cases show that base (blue boxes) and novel classes (red boxes) are correctly detected in one image. In the bottom row, there are some misclassified results for failure cases due to similar characteristics and appearance. From left to right: the boat is misclassified as an airplane, the birds are confused for sheep and cow, and the train is detected well, but the bus is detected even though there is no bus.
%  These qualitative results show the applicability of our method to real-world problems. 
\begin{table}[hbt!]
\renewcommand{\arraystretch}{1.1}
\centering
\caption{Performance of the cross-domain experiment from COCO to PASCAL VOC.} \label{tab:tab4}
{\footnotesize
\resizebox{\columnwidth}{!}{%
\begin{tabular}{c|ccccc}
\Xhline{3\arrayrulewidth}
\multirow{2}{*}{Shots} & \multicolumn{5}{c}{Method}           \\ \cline{2-6} 
                      & Ours & DBF \cite{li2021towards}  & LEAST \cite{li2021class} & LEAST-NE \cite{li2021class} & ONCE \cite{perez2020incremental}\\ \Xhline{2\arrayrulewidth}
1                     & \textbf{11.3} & -    & 6.8   & \underline{7.2}      & -    \\
5                     & \textbf{26.0} & 12.5 & \underline{16.7}  & 16.5     & 2.4  \\
10                    & \textbf{30.4} & \underline{23.7} & 18.8  & 18.5     & 2.6  \\ \Xhline{3\arrayrulewidth}
\end{tabular}%
}}
\end{table}

\section{CONCLUSIONS}

In this paper, we proposed iTFA for iFSD. To solve iFSD, we addressed a simple fine-tuning strategy by modifying TFA. We divided the RoI feature extractor and classifier into the base and novel branches and then fine-tuned only the novel branch. We also adopted a class-agnostic regressor and analyzed the compatibility of the classifier. Our method outperformed the prior meta-learning methods by a large margin in widely used datasets. Although our method fine-tuned the model without exemplars of the base classes, it showed competitive results with other fine-tuning-based methods. We will adapt our fine-tuning strategy to various object detector structures for further improvement.

% \addtolength{\textheight}{-12cm}   % This command serves to balance the column lengths
                                  % on the last page of the document manually. It shortens
                                  % the textheight of the last page by a suitable amount.
                                  % This command does not take effect until the next page
                                  % so it should come on the page before the last. Make
                                  % sure that you do not shorten the textheight too much.

%%%%%%%%%%%%%%%%%%%%%%%%%%%%%%%%%%%%%%%%%%%%%%%%%%%%%%%%%%%%%%%%%%%%%%%%%%%%%%%%

%%%%%%%%%%%%%%%%%%%%%%%%%%%%%%%%%%%%%%%%%%%%%%%%%%%%%%%%%%%%%%%%%%%%%%%%%%%%%%%%

%%%%%%%%%%%%%%%%%%%%%%%%%%%%%%%%%%%%%%%%%%%%%%%%%%%%%%%%%%%%%%%%%%%%%%%%%%%%%%%%
% \section*{APPENDIX}

% \section*{ACKNOWLEDGMENT}

%%%%%%%%%%%%%%%%%%%%%%%%%%%%%%%%%%%%%%%%%%%%%%%%%%%%%%%%%%%%%%%%%%%%%%%%%%%%%%%%

% \begin{thebibliography}{99}
% \bibliography{ref}
% \end{thebibliography}
\bibliographystyle{IEEEtran}
\bibliography{ref}

\end{document}